\documentclass[runningheads]{llncs}
\usepackage[T1]{fontenc}
\usepackage{graphicx}
\usepackage[table,x11names]{xcolor}
\usepackage{enumitem}
\usepackage{algorithm}
\usepackage{algpseudocode}
\usepackage{amsmath}
\usepackage{amssymb}
\usepackage{subcaption}
\usepackage{float}
\usepackage{wrapfig}
\usepackage{placeins}
\usepackage{graphicx}
\usepackage{xcolor}
\usepackage[colorlinks=true, linkcolor=blue, urlcolor=blue, citecolor=blue]{hyperref}
\usepackage{array}
\usepackage{multirow}
\usepackage{orcidlink}
\usepackage{marvosym}
\newcommand{\taug}{\ensuremath{\tau_{\mathrm{grad}}}}
\newcommand{\taur}{\ensuremath{\tau_{\mathrm{ratio}}}}
\newcolumntype{C}[1]{>{\centering\arraybackslash}p{#1}}
\begin{document}
\title{Extreme Views: 3DGS Filter for Novel View Synthesis from Out-of-Distribution Camera Poses}
\titlerunning{EV3DGS}
\author{Damian Bowness$^{(\text{\Letter})}$\orcidlink{0009-0008-2245-3525} \and Charalambos Poullis\orcidlink{0000-0001-5666-5026}}
\authorrunning{D. Bowness et al.}

\institute{Concordia University, Montréal, Québec, Canada\\
\email{damian.bowness@gmail.com, charalambos@poullis.org}}

\maketitle              
\begin{abstract}
When viewing a 3D Gaussian Splatting (3DGS) model from camera positions significantly outside the training data distribution, substantial visual noise commonly occurs. These artifacts result from the lack of training data in these extrapolated regions, leading to uncertain density, color, and geometry predictions from the model.\\

To address this issue, we propose a novel real-time render-aware filtering method. Our approach leverages sensitivity scores derived from intermediate gradients, explicitly targeting instabilities caused by anisotropic orientations rather than isotropic variance. This filtering method directly addresses the core issue of generative uncertainty, allowing 3D reconstruction systems to maintain high visual fidelity even when users freely navigate outside the original training viewpoints.\\

Experimental evaluation demonstrates that our method substantially improves visual quality, realism, and consistency compared to existing Neural Radiance Field (NeRF)-based approaches such as BayesRays. Critically, our filter seamlessly integrates into existing 3DGS rendering pipelines in real-time, unlike methods that require extensive post-hoc retraining or fine-tuning.\\

Code and results at \href{https://damian-bowness.github.io/EV3DGS}{https://damian-bowness.github.io/EV3DGS}

\end{abstract}
\section{Introduction}
High-fidelity 3D scene reconstruction and novel view synthesis represent fundamental challenges in computer vision and graphics, with applications spanning virtual reality, autonomous navigation, and digital content creation. The quality of rendered novel views directly depends on the completeness and accuracy of the underlying 3D scene representation, making robust reconstruction from limited viewpoints a critical research priority.

Recent advances in neural 3D reconstruction, particularly Neural Radiance Fields (NeRF) and 3D Gaussian Splatting (3DGS), demonstrate remarkable capabilities for photorealistic novel view synthesis. 3DGS offers particular advantages for real-time applications due to its explicit 3D primitive representation and efficient rasterization-based rendering pipeline.

However, a fundamental limitation emerges when these reconstructed 3D models are rendered from camera viewpoints significantly outside the original training distribution. The 3D reconstruction process lacks sufficient multi-view constraints to accurately estimate scene geometry and appearance in extrapolated regions. This manifests as rendering artifacts including floating primitives, inconsistent geometry, and view-dependent noise that severely compromise visual quality.

\begin{wrapfigure}{r}{0.45\linewidth} 
  \vspace{-\baselineskip}             
    \centering
    \includegraphics[width=.95\linewidth]{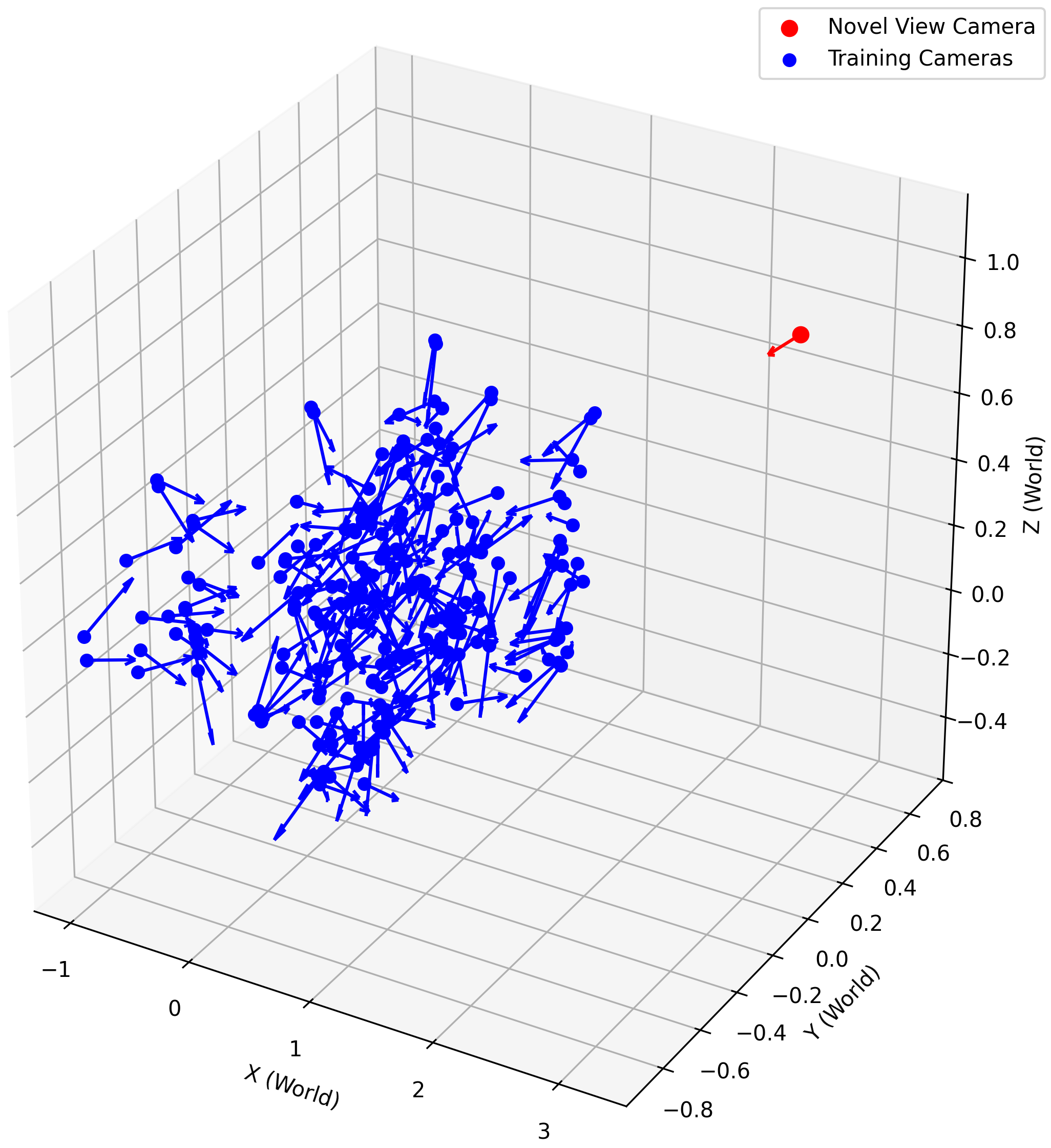}
    \caption{Novel view synthesis from an OOD camera pose(red) compared to the training cameras poses (blue).}
    \label{fig:ood}
    \vspace{-\baselineskip}   
\end{wrapfigure}
During training, 3DGS optimizes Gaussian primitive parameters using photometric supervision from available camera views. However, regions visible from extreme out-of-distribution (OOD) viewpoints (see Fig.~\ref{fig:ood}) often lack adequate multi-view coverage, leading to poorly-constrained optimization. This results in Gaussian primitives with uncertain 3D properties that produce artifacts when rendered from novel camera poses.

Existing solutions attempt to address these issues through improved training strategies, additional regularization, or uncertainty quantification during reconstruction. However, such approaches typically require expensive retraining procedures or fundamental modifications to the 3D reconstruction pipeline, making them impractical for deployment scenarios where pre-trained models must accommodate viewing trajectories that differ significantly from the original data collection conditions.

We propose a fundamentally different approach: Rather than modifying the 3D reconstruction process itself, we introduce a real-time filtering method that operates during rendering to identify and suppress unstable 3D primitives. Our key insight is that rendering artifacts from out-of-distribution views primarily result from anisotropic Gaussian primitives with elongated shapes or poor orientations that reflect insufficient multi-view constraints during training.

Our method computes gradient-based sensitivity measures that capture how rendered pixel colors respond to perturbations in 3D space. By analyzing these gradients in a rotation-aligned coordinate system, we can identify Gaussian primitives whose 3D properties lead to view-dependent instabilities. This enables targeted removal of problematic primitives during rendering to improve novel view synthesis quality.

The resulting system seamlessly integrates into existing 3DGS rendering pipelines, requiring no modifications to the underlying 3D reconstruction method or additional training data. This enables robust rendering from extreme viewpoints while preserving real-time performance.  Our technical contributions include:
\begin{itemize}
    \item A novel gradient-based sensitivity analysis specifically designed to identify unstable 3D Gaussian primitives resulting from incomplete multi-view reconstruction 
    \item A rotation-aligned filtering approach that targets anisotropic reconstruction uncertainties without requiring 3D model retraining 
    \item Comprehensive validation demonstrating significant improvements in novel view synthesis quality from OOD camera poses 
\end{itemize}
\section{Related Work}
\label{sec:related_work}
\subsection{Neural Radiance Fields}
Neural rendering techniques, particularly Neural Radiance Fields (NeRF)~\cite{nerf}, have significantly advanced novel-view synthesis by implicitly representing 3D scenes as continuous volumetric fields encoded via neural networks. NeRF optimizes a multi-layer perceptron (MLP) by sampling along camera rays to predict color and opacity values, enabling photorealistic image rendering from unseen viewpoints. However, this implicit representation poses challenges such as high computational cost and slow rendering times, limiting their real-time application potential.

To address uncertainties in NeRF reconstructions, Goli et al. ~\cite{bayesrays} proposed BayesRays, a Bayesian framework using Laplace approximation to quantify systematic uncertainties. However, BayesRays requires a post-hoc training procedure to build uncertainty fields, making it impractical for deployment scenarios where models must be used immediately after initial reconstruction without additional optimization phases.

\subsection{3D Gaussian Splatting}
Recent efforts shifted toward explicit volumetric representations such as Plenoxels~\cite{plenoxels} and 3DGS~\cite{3dgs}, addressing computational inefficiencies and enabling real-time rendering. Specifically, 3DGS uses a set of explicit Gaussian primitives to represent radiance fields efficiently.

Building on explicit representations, several works have explored uncertainty quantification and rendering improvements for 3DGS.  Jiang et al.~\cite{fisherrf} generalize uncertainty quantification using Fisher information-based approximation across training images, targeting optimal view selection during training but not addressing rendering-time instabilities or anisotropic uncertainties.  Hanson et al.~\cite{pups}  integrate uncertainty attributes into 3DGS for pruning redundant primitives, though their approach primarily targets spatial redundancy rather than artifacts from directional instabilities during novel view rendering.

Methods exploiting ray-Gaussian intersections have emerged to improve rendering accuracy and facilitate geometry extraction. Keselman and Hebert~\cite{meatballs} develop a differentiable ray-based renderer that integrates algebraic surfaces with Gaussian mixture models (GMMs). Their work provides analytic solutions for computing intersections between rays and Gaussian primitives.  Extending these concepts, Gao et al.~\cite{relight} adapt ray-Gaussian intersection solutions for 3DGS to improve lighting effects and rendering. Yu et al.~\cite{gof} construct volumetric opacity fields from ray-Gaussian intersections for high-quality mesh extraction.

Despite significant advances, existing uncertainty quantification methods primarily target isotropic noise, redundant primitives, or implicit representations. There is a notable gap in efficiently addressing anisotropic instabilities and directional sensitivity inherent in explicit representations such as 3DGS. Current approaches either incur substantial computational overhead, require retraining, or inadequately address directional instabilities that significantly degrade rendering quality and temporal coherence.
\section{Background}
\label{sec:background}
3DGS is a hybrid volumetric rendering technique that models radiance fields using a discrete set of explicit Gaussian volumes. By representing radiance fields as a set of 3D Gaussians, 3DGS defines the probability distribution of the radiance of a point in space. Each 3D Gaussian primitive, $\mathcal{G}$, is independently parameterized by its density parameters and spatial parameters. The density parameters are color, $c$, and opacity, $\alpha$.  The spatial parameters are the mean, $\mu$, and covariance, $\Sigma$.

\begin{align}
    \mathcal{G}(x) = e^{-\tfrac{1}{2}(x-\mu)^T\Sigma^{-1}(x-\mu)}
\end{align}

To ensure the covariance matrix remains positive semi-definite throughout optimization the 3D Gaussians are represented as ellipsoids.  Therefore the Gaussian mean is represented by the ellipsoid's center while the covariance matrix is decomposed into a scaling matrix, $S$, and rotation matrix, $R$,

\begin{align}
   \Sigma = RSS^TR^T
\end{align}

For rendering, 3D Gaussians are depth-sorted and projected onto a 2D image plane.  The color, $C$, of pixel, $x$, is computed by alpha compositing:

\begin{align}\label{eq:2}
    C(x) = \sum\limits_{i=1}^N c_i\alpha_i \mathcal{G}_i(x_i)\bigg[\prod\limits_{j=1}^{i-1} \big(1-\alpha_j\mathcal{G}_j(x_j)\big)\bigg]
\end{align}

The parameters of each 3D Gaussian are optimized from back propagating the loss between the rendered image and ground-truth image.

\section{Methodology}
\label{sec:methodology}
We introduce a render-time gradient sensitivity analysis for 3D Gaussian Splatting. For every Gaussian that a camera ray intersects, we find the depth along the ray where that Gaussian contributes most to the pixel.  We then assign a sensitivity score indicating how much the pixel color would change under tiny spatial nudges of position, orientation, or scale.  Low scores mark stable and trustworthy contributions; high scores flag regions likely to produce artifacts. To better detect directional instabilities, the score is evaluated in a rotation-aligned coordinate system that emphasizes orientation effects without over-penalizing fine detail.

\begin{figure}[H]
    \centering
    \includegraphics[width=.95\linewidth]{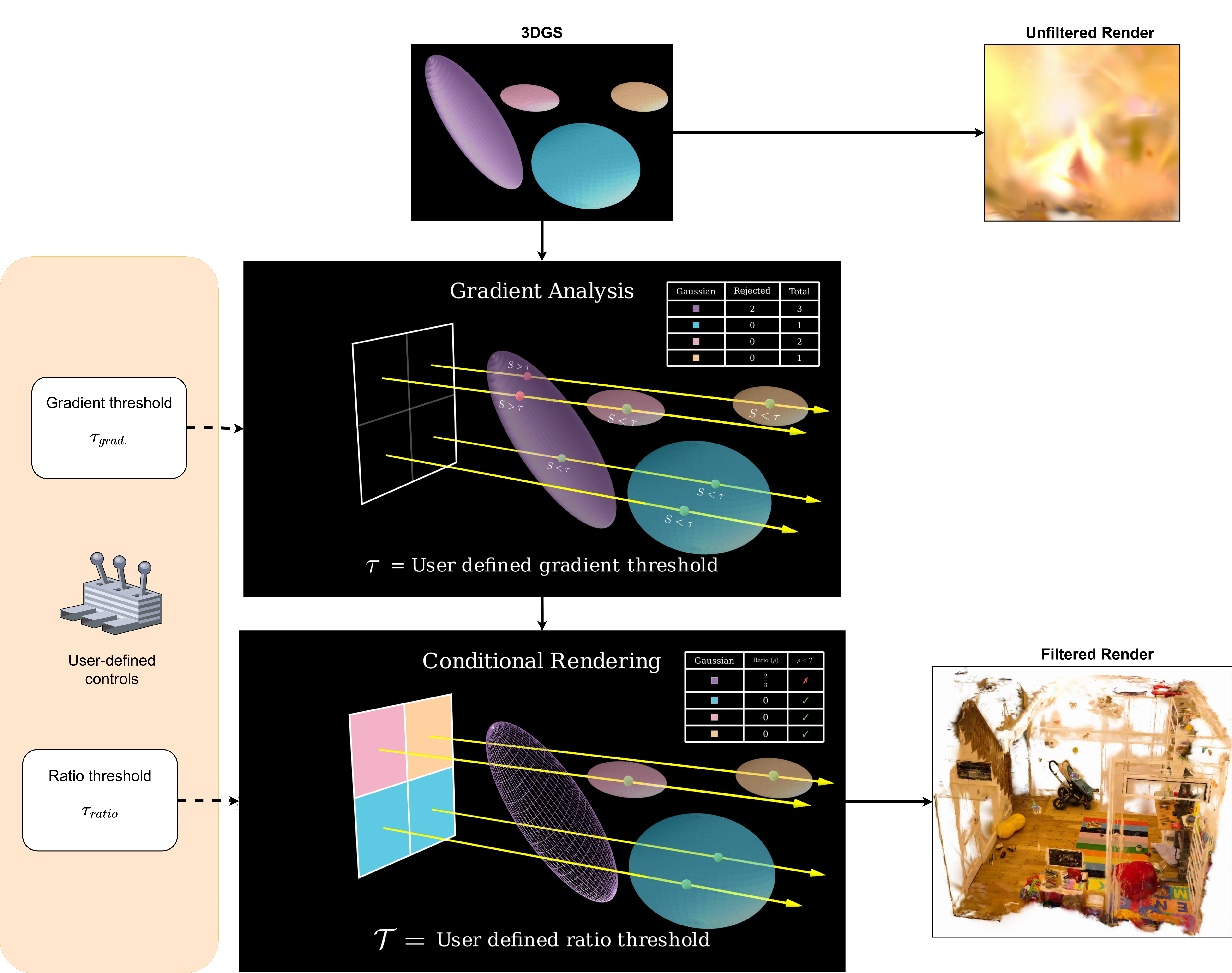}
    \caption{Pipeline of our two-pass filter}
    \label{fig:pipeline}
\end{figure}

For a given view, we use a two-pass approach to select which Gaussians are used to render the view (see Fig.~\ref{fig:pipeline}).  In the first pass, each intersection is marked accepted or rejected based on the sensitivity threshold, and for every Gaussian we count both rejected and total intersections. In the second pass, we compute the rejection fraction for each Gaussian; if it exceeds a user-set limit, that Gaussian is removed from the rendering process for the current viewpoint.

This targeted, view-conditioned filtering preserves stable, detail-carrying Gaussian primitives while suppressing unstable ones, reducing flicker and floaters and improving visual quality and geometric consistency under extreme OOD camera poses.
\vspace{-0.5em}

\subsection{Ray-Marching}

To analyze the radiance field with fine spatial resolution, we adopt a ray-marching approach. Unlike screen-space projection methods that evaluate Gaussians in 2D after rasterization, ray-marching enables direct computation of the ray-Gaussian interaction in 3D space, leading to more precise control over rendering dynamics. This method is particularly suited for computing pointwise sensitivity, as it allows us to localize the analysis to specific ray-Gaussian intersections.

Given a ray defined by its origin $o \in \mathbb{R}^3$ and direction $r \in \mathbb{R}^3$, any point $x$ on the ray is parameterized as:
\begin{equation}
    x = o + t r
\end{equation}
where $t$ represents the distance along the ray.

To evaluate the contribution of a $k^{th}$ 3D Gaussian to the ray, we first transform the ray into the Gaussian’s canonical coordinate system. This transformation normalizes spatial variation using the Gaussian’s scale $S_{k}$ and orientation $R_{k}$, and positions the ray relative to the Gaussian’s mean $\mu_{k}$:

\begin{align}
    \label{eq:og} o_g &= S_{k}^{-1} R_{k} (o - \mu_{k})\\
    r_g &= S_{k}^{-1} R_k r\\
   \label{eq:xg} x_g &= o_g + t r_g 
\end{align}

In this normalized space, the Gaussian contribution simplifies to a 1D univariate Gaussian along the ray:

\begin{align}
    G^{1D}(t) &= e^{-\frac{1}{2} x_g^T x_g}\\
    &= e^{-\frac{1}{2} (r_g^T r_g t^2 + 2 o_g^T r_g t + o_g^T o_g)}
\end{align}

This quadratic form of the exponent provides analytical tractability and numerical stability, facilitating efficient determination of the maximum Gaussian contribution without needing to solve higher-order equations.

The peak contribution occurs at the depth $t_{\text{min}}$, where the exponent reaches its minimum:

\begin{equation}
    t_{\text{min}} = -\frac{o_g^T r_g}{r_g^T r_g}
\end{equation}

After determining the depths for all Gaussians intersected by the ray, the final pixel color is computed via depth-ordered alpha compositing across all Gaussians intersected by the ray:

\begin{align}
    C(o,r) = \sum_{k=1}^K c_k \alpha_k \mathcal{G}_k^{1D}(t_{k,\text{min}}) \prod_{j=1}^{k-1}(1 - \alpha_j \mathcal{G}_j^{1D}(t_{j,\text{min}}))
\end{align}

This formulation allows precise control over how Gaussians influence each pixel, by ensuring accurate modeling of cumulative optical effects along viewing rays, laying the groundwork for analyzing how sensitive each pixel is to local changes in the 3D radiance field.

\subsection{Color Gradient Sensitivity}

To measure the sensitivity of the rendered color to spatial perturbations, we derive the gradient of the composite color $C(o, r)$ with respect to the 3D position $x$. This involves computing the gradient of the alpha-blended color contribution from each Gaussian, taking into account how both direct and accumulated transmittance change under spatial variation.

Starting from the definition, we express the gradient of the composite color as:

\begin{align}\label{eq:3}
    \nabla C(x) = \sum_{k=1}^K c_k a_k \nabla \prod_{j=1}^{k-1} (1-a_j) + \prod_{j=1}^{k-1}(1-a_j) \nabla \mathcal{G}_k(x_k)
\end{align}
where $a_i = \alpha_i \mathcal{G}_i(x_i)$. The gradient of a single Gaussian term is given by:

\begin{align}
    \nabla_x \mathcal{G}(x) = e^{-\tfrac{1}{2}x^T\Sigma^{-1} x}(-\tfrac{1}{2})\big(2\Sigma^{-1} x\big) = - \mathcal{G}(x) \Sigma^{-1} x
\end{align}
and the gradient of the accumulated transmittance product becomes:

\begin{align}
    \nabla_x \prod_j (1 - a_j) = \left( \prod_j (1 - a_j) \right) \sum_j \frac{a_j \Sigma_j^{-1} x_j}{1 - a_j}
\end{align}

Substituting these expressions into the original gradient equation yields:

\begin{align}\label{eq:jacobish}
    \nabla C = \sum_{k=1}^K c_k a_k \prod_{j=1}^{k-1}(1 - a_j) \left( \sum_{j=1}^{k-1} \frac{a_j \Sigma_j^{-1} x_j}{1 - a_j} - \Sigma_k^{-1} x_k \right)
\end{align}

This Jacobian-like quantity describes how sensitive the final color is to the underlying spatial configuration. However, due to the involvement of matrix inversions and eigenvalue computations, evaluating this fully is computationally expensive and impractical for real-time rendering.

To improve computational practicality, we decouple the gradient from the color vector $c_k$ by replacing it with the scalar constant $c_k = 1$. This removes color-specific variation and focuses on transmittance dynamics. This form captures the structural sensitivity of the scene to spatial changes, identifying regions of high rendering instability while avoiding unnecessary per-color calculations.

\subsection{Directional Sensitivity and Rotation-Based Gradient Filtering}\label{sec43}

While scalar gradient magnitudes effectively quantify overall spatial sensitivity, they do not capture directional instabilities.  That is, situations in which the rendered output is disproportionately sensitive to perturbations along specific directions or axes. To address this limitation, we extend our analysis by calculating directional gradients within a rotation-aligned coordinate system. By isolating the influence of Gaussian orientation from its scale, this method enables a precise assessment of sensitivity relative to the Gaussian's principal axes. Consequently, we can independently evaluate directional instabilities without the confounding effects introduced by anisotropic scaling.

The core intuition is that Gaussians with strong anisotropic properties (i.e. those elongated along 1 or 2 of the 3 axes) exhibit direction-dependent instability. For example, a long, thin Gaussian may be stable along its major axis but highly sensitive to perturbations along its minor axes. Traditional geometric analyses such as Principal Component Analysis (PCA) use eigenvalue ratios of the covariance matrix to characterize such anisotropy, but these are not render-aware and they do not directly measure the impact on rendered output.

Instead, by applying the gradient filter in a rotation-only space, we can expose directional rendering instabilities. This is achieved by transforming the covariance matrix using only its rotation matrix $R_k$, excluding the scale $S_k$. In this aligned space, we compute sensitivity gradients relative to each principal axis of the Gaussian, revealing how rendering behavior changes with orientation.  Therefore, we drop the scale transformation in Equations \eqref{eq:og}–\eqref{eq:xg} and derive our sensitivity metric: 
\begin{align}\label{eq:grad}
    S = \sum_{k=1}^K a_k \prod_{j=1}^{k-1}(1 - a_j) \left( \sum_{j=1}^{k-1} \frac{a_j x_j}{1 - a_j} - x_k \right)
\end{align}
where $x = R_k(o - \mu_k) + t R_kr$, which represent the ray-Gaussian intersection point in the rotation-aligned Gaussian space.

This approach offers several advantages:
\begin{itemize}
    \item \textbf{Isolates rotational sensitivity}: By excluding scale, we ensure that the gradient reflects only changes due to orientation, not magnitude.
    \item \textbf{Highlights unstable orientations}: High directional sensitivity indicates that minor misalignments can significantly affect rendering, flagging Gaussians prone to producing visual artifacts.
    \item \textbf{Complementary to PCA filtering}: While PCA identifies noise due to isotropic variations, our method detects and corrects instabilities due to anisotropic orientation.
\end{itemize}

\subsection{Aggregate Sensitivity Analysis}

We introduce a two-pass filtering pipeline to evaluate and reject unstable Gaussians based on their aggregate sensitivity scores.

In the first pass, we compute gradient-based sensitivity at each ray-Gaussian intersection, as defined in Equation~\ref{eq:grad}. A Gaussian's contribution to a pixel's color is conditionally accepted or rejected according to a user-defined sensitivity threshold ($\tau_{grad}$). For each Gaussian, a ray intersection's contribution is either accepted or rejected.  We track 2 counts: rejected and total, where total is the sum of accepted and rejected counts. These counts serve as inputs for the second pass.

In the second pass, we compute the aggregate sensitivity score for each Gaussian as the ratio of its rejected count to its total usage count. The aggregate sensitivity score for a Gaussian with a zero usage count is 1.  Gaussians with a rejection ratio exceeding a user-defined threshold ($\tau_{ratio}$) are excluded from the final rendering.

Ultimately, this filtering mechanism enables robust scene reconstruction by attenuating the influence of unstable Gaussians. By selectively removing Gaussians with high aggregate sensitivity, we reduce noise and enhance the spatial consistency of the rendered view. This targeted use of rotation-aligned gradient analysis allows us to remove distractors and improve rendering quality.
\section{Experiments and Results}
\label{sec:results}
\subsubsection{Experimental setup.} We evaluate a modified Nerfstudio Splatfacto pipeline \cite{nerfstudio}, replacing the default projection with ray marching and inserting our gradient-sensitivity computation plus two-pass filter for extreme OOD views. We use fixed thresholds: $\taug$=0.0001 and $\taur$=0.5, except for the Doctor Johnson scene from the Deep Blending dataset which uses $\taur$=0.01.  Threshold values are chosen heuristically to suppress extreme, instability-inducing sensitivities.

Splatfacto models use COLMAP initialization and 30k iterations. Following BayesRays, we train a Nerfacto model for 30k iterations and extract uncertainty fields over an additional 1k iterations. Experiments are run on an RTX 2080 Ti; images are downscaled by 30–50\% ($\approx$ 1500×700 px) for runtime constraints.
\subsubsection{Datasets \& protocol.} We test on Deep Blending and NeRF On-the-go \cite{deepblending,nerfonthego}. For Deep Blending, we render extreme OOD views by extrapolating far beyond the training poses. For NeRF On-the-go, we render along the original trajectories and compare to ground truth to gauge suppression of transient artifacts. Perceptual quality is measured with no-reference image quality (NR-IQA) metrics: NIQE, BRISQUE, and PIQE.

\begin{itemize}[itemsep=0pt, parsep=0pt, topsep=0pt]
    \item \textbf{Natural Image Quality Evaluator (NIQE)}:\\
    assesses image quality based on deviations from a learned natural scene statistics model.~\cite{niqe}
    \item \textbf{Blind/Referenceless Image Spatial Quality Evaluator (BRISQUE)}:\\
    uses statistical features from locally normalized luminance coefficients to predict perceived distortion. ~\cite{brisqe}
    \item \textbf{Perception based Image Quality Evaluator (PIQE)}:\\
    quantifies perceptual distortions by analyzing block-wise degradation in an image, emphasizing spatially significant distortions.~\cite{piqe}
\end{itemize}
These metrics are designed to capture human-perceived image quality by penalizing unnatural textures, noise, and distortions in the absence of reference images.   Extreme OOD views can produce large white regions, which bias NR-IQA; therefore all renderings are cropped prior to scoring.  For all three metrics, lower values are better. 
\subsubsection{Evaluation.} For each scene we render an animation that moves from in-distribution viewpoints to an OOD orbital path. We compute NIQE, BRISQUE, PIQE per frame and report per-scene averages in Table~\ref{table:results}.  To our knowledge our method is the first real-time filter for 3DGS.  Therefore we compare to BayesRays \cite{bayesrays}, a real-time NeRF filtering baseline using a pre-trained uncertainty field.  

\begin{figure}[!ht]
\raggedleft
\setlength{\tabcolsep}{8pt}      
\renewcommand{\arraystretch}{1.0}
\begin{tabular}{cc|cc}
\multicolumn{2}{c|}{\textbf{Playroom}} & \multicolumn{2}{c}{\textbf{Creepy Attic}}\\ \hline
\textbf{Unfiltered} & \textbf{Filtered} & \textbf{Unfiltered} & \textbf{Filtered} \\
\includegraphics[width=.2\textwidth]{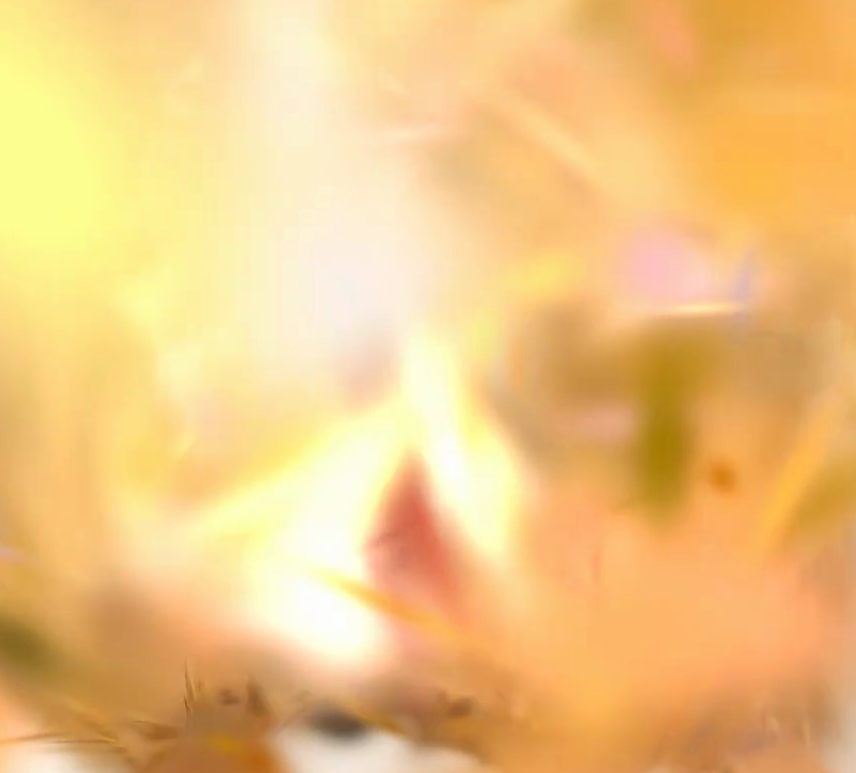} &
\includegraphics[width=.2\textwidth]{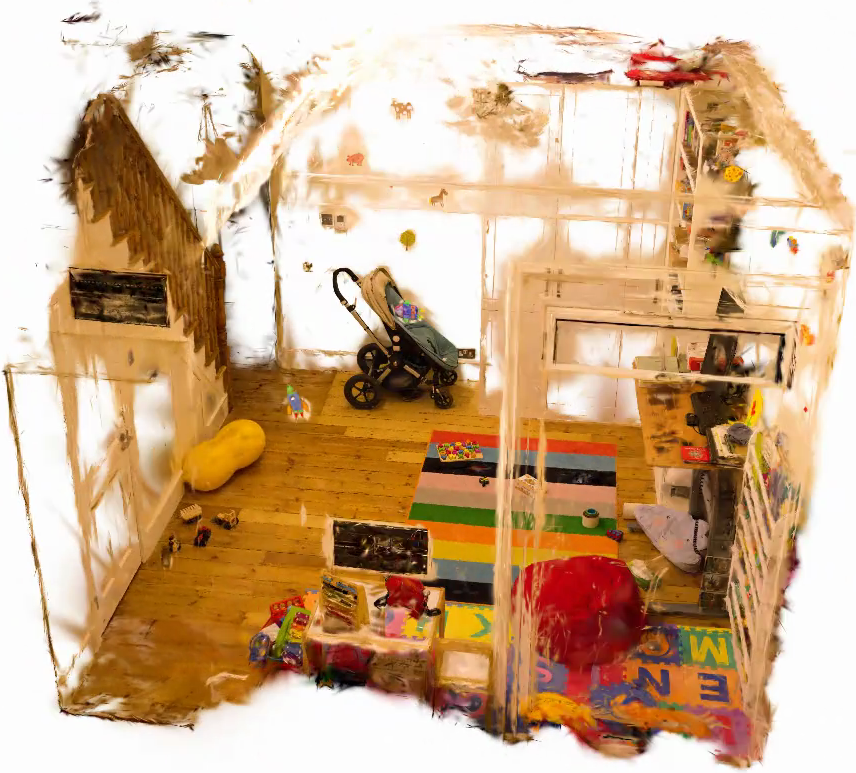} &
\includegraphics[width=.27\textwidth] {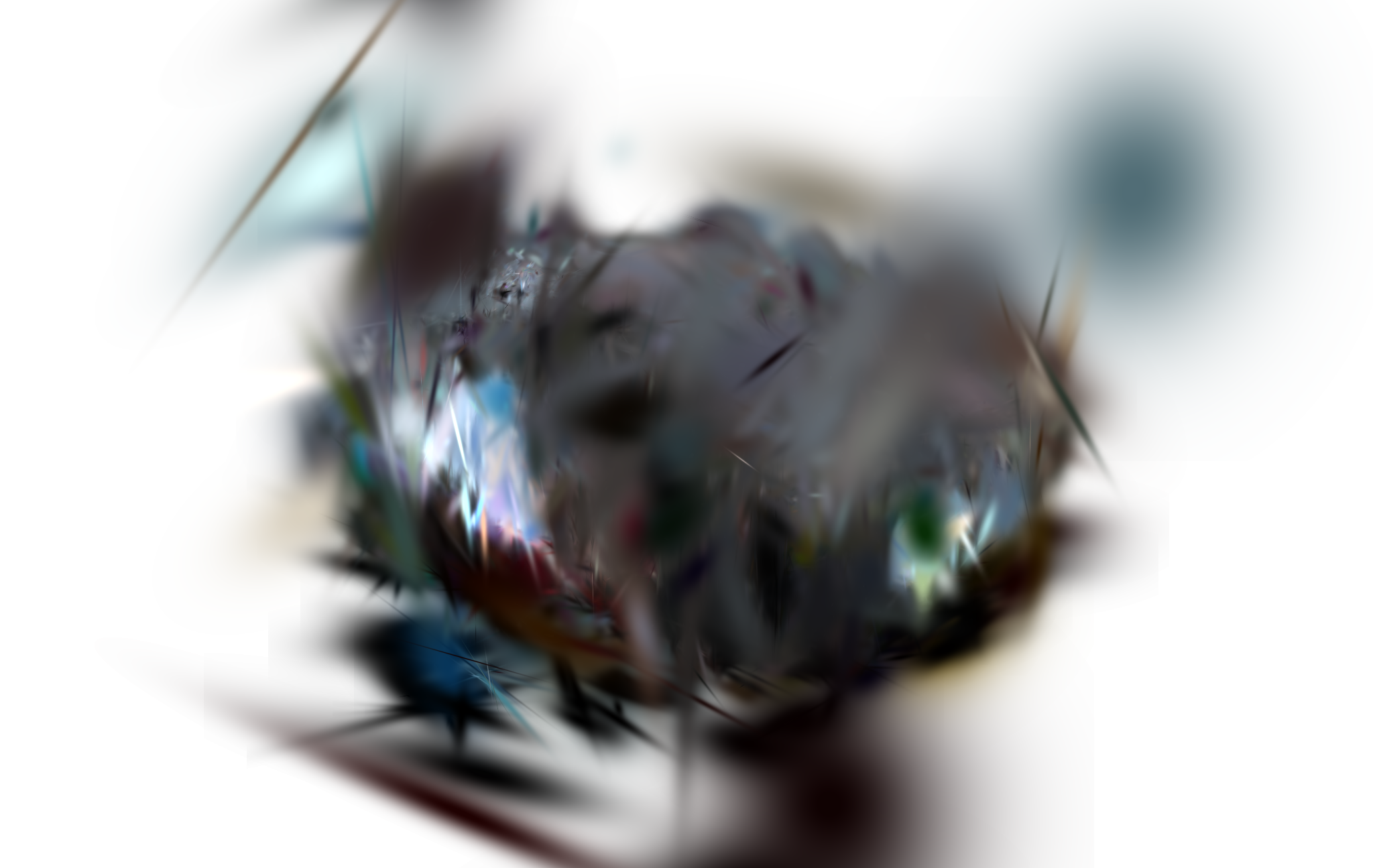} &
\includegraphics[width=.27\textwidth]{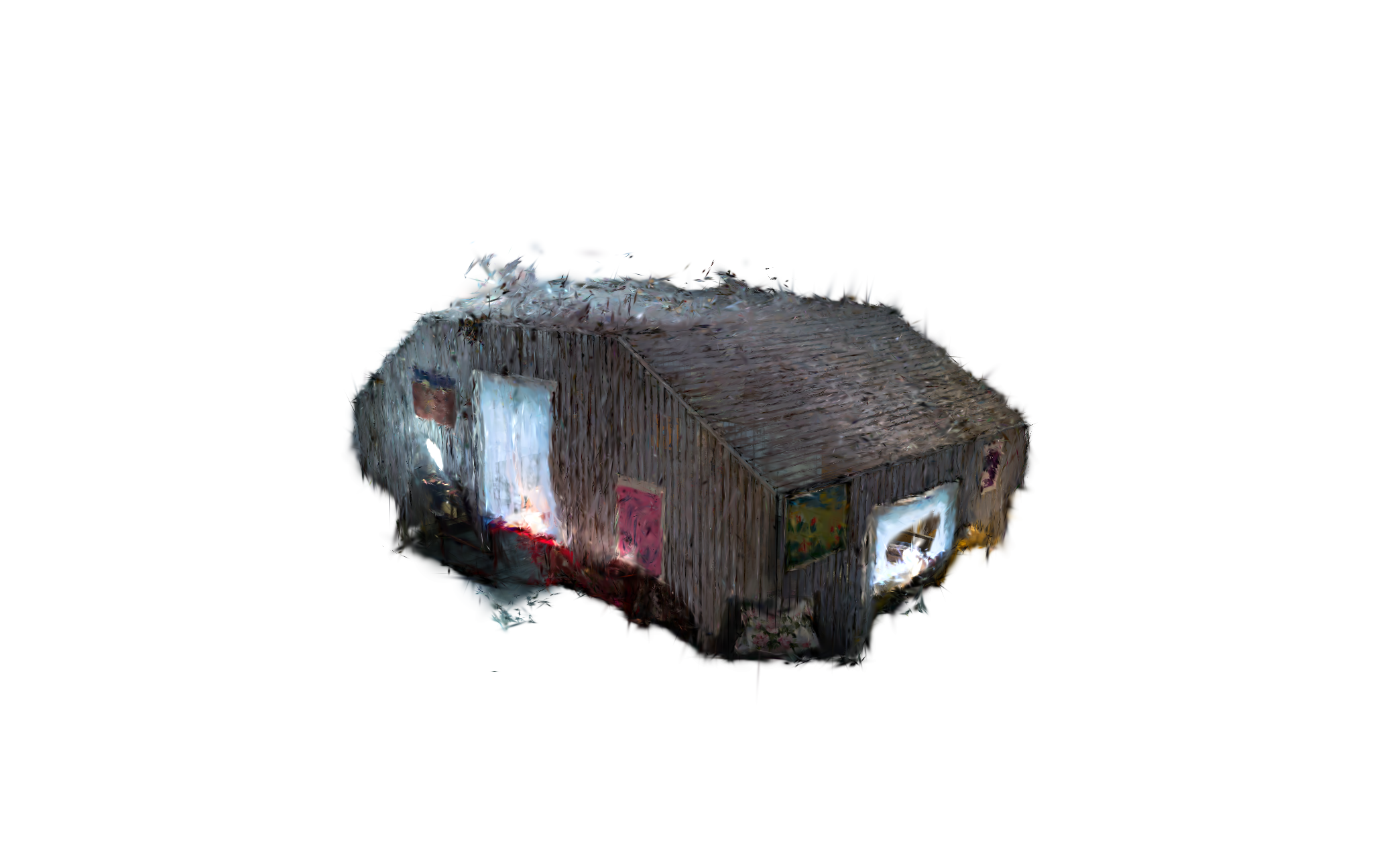}
\end{tabular}\\
\vspace{1em}
\begin{tabular}{cc}
 \multicolumn{2}{c}{\textbf{Dr. Johnson}}\\ \hline
 \textbf{Unfiltered} & \textbf{Filtered} \\
\includegraphics[width=.5\textwidth]{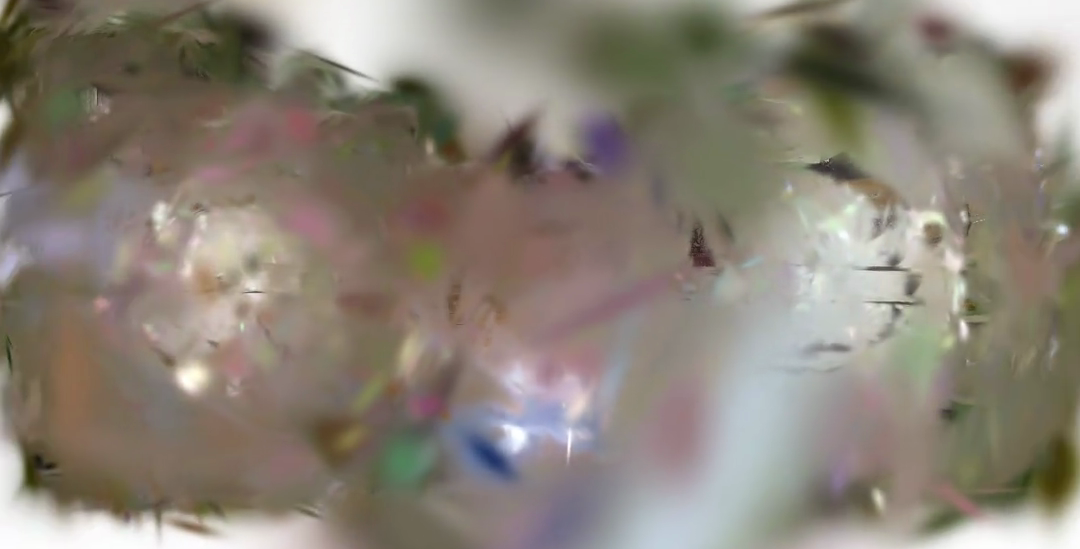} &
\includegraphics[width=.5\textwidth]{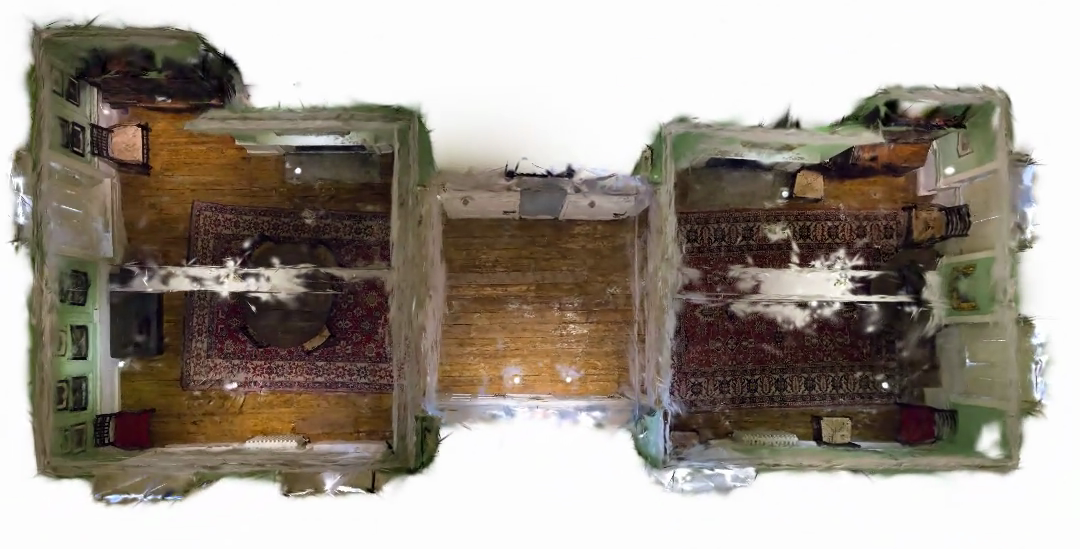}
\end{tabular}
\caption{Deep Blending example frames.}
\label{fig:example-frames}
\end{figure}

Our filter achieves the lowest (best) NIQE, BRISQUE, and PIQE across all scenes (see table~\ref{table:results}). These gains in perceptual quality across extreme OOD camera positions indicate effective suppression of anisotropy-induced artifacts without over-smoothing.  Example frames (unfiltered vs. filtered) for each scene are provided in Fig.~\ref{fig:example-frames}.
\begin{table}[!h]
\centering
\begin{tabular}{|l|C{1cm}|C{1cm}|C{1cm}|C{1cm}|C{1cm}|C{1cm}|C{1cm}|C{1cm}|C{1cm}|}
\hline
     & \multicolumn{3}{c|}{Playroom} & \multicolumn{3}{c|}{Creepy Attic} & \multicolumn{3}{c|}{Dr. Johnson}\\ \hline
    \rowcolor{gray!20} & \textbf{NQ} $\downarrow$ & \textbf{BR} $\downarrow$ & \textbf{PQ} $\downarrow$ & \textbf{NQ} $\downarrow$ & \textbf{BR} $\downarrow$ & \textbf{PQ}  $\downarrow$ & \textbf{NQ} $\downarrow$ & \textbf{BR} $\downarrow$ & \textbf{PQ}  $\downarrow$\\ \hline
    BayesRays 0.1 & 11.74 & 44.45 & 83.84 & 10.30 & 45.26 & 77.42 & 10.02 & 46.84 & 70.38 \\ \hline
    BayesRays 0.2 & 9.18  & 47.11 & 74.47 & 10.17  & 55.12 & 69.64 & 9.71  & 57.88 & 59.58 \\ \hline
    BayesRays 0.5 & 10.67 & 64.24 & 58.31 & 14.02 & 53.57 & 46.87 & 11.34 & 55.91 & 59.76  \\ \hline
    \rowcolor{yellow} \textbf{Ours} & \textbf{3.41}  & \textbf{41.38} & \textbf{52.99}  & \textbf{5.94}  & \textbf{42.30} & \textbf{40.24}  & \textbf{4.49}  & \textbf{37.37} & \textbf{51.60} \\ \hline
\end{tabular}
\caption{NIQE (NQ), BRISQUE (BR), and PIQE (PQ) scores for Playroom, CreepyAttic, and DrJohnson.}
\label{table:results}
\end{table} 

\subsubsection{Connection to information measures.} 
\begin{figure}[!h]
  \begin{tabular}{cccc}
    \centering
    & \textbf{Original} & \textbf{3DGS} & \textbf{Our Filter}\\
    arcdetriomphe &
    \includegraphics[width=.3\linewidth]{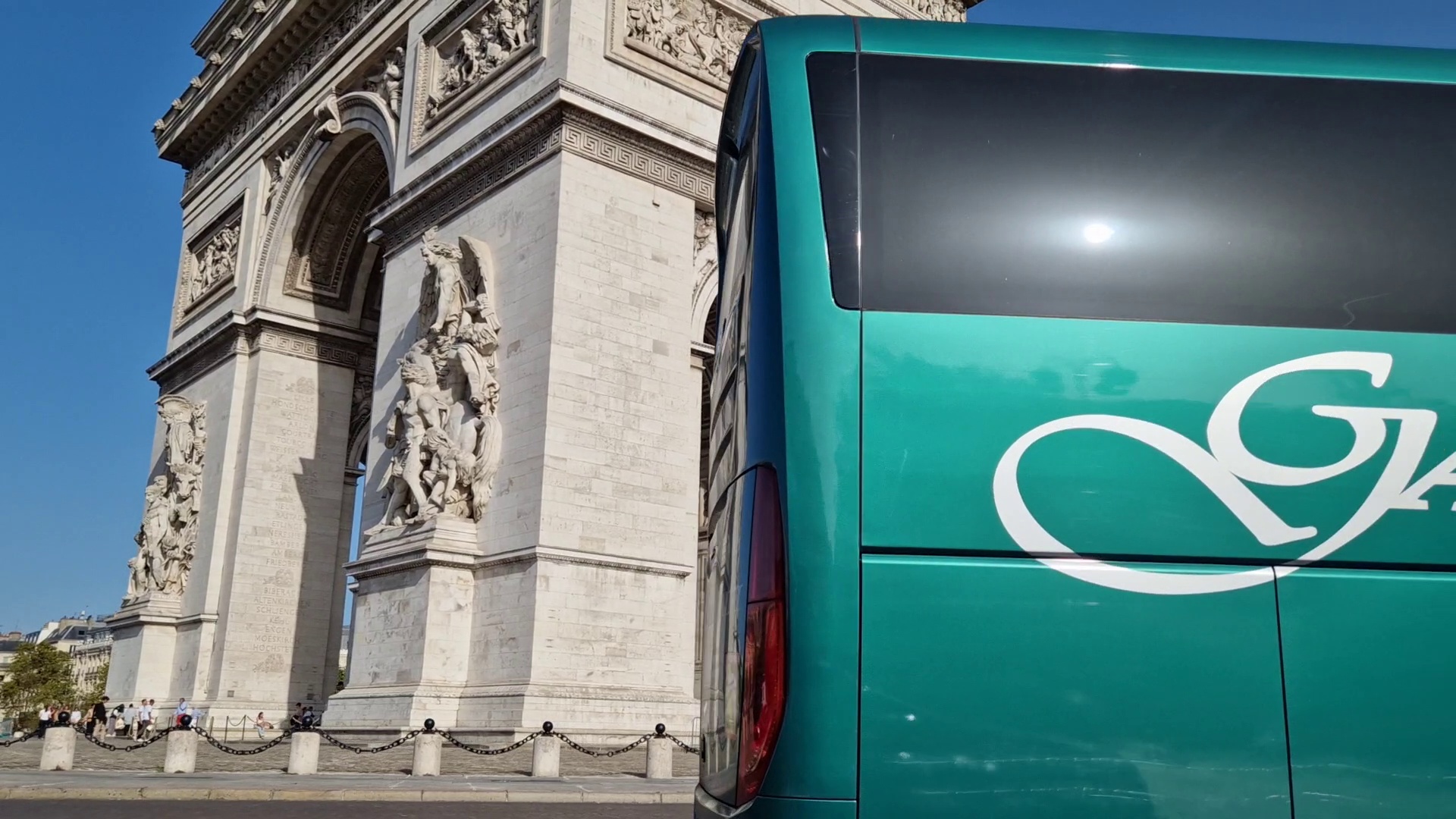}&
    \includegraphics[width=.3\linewidth]{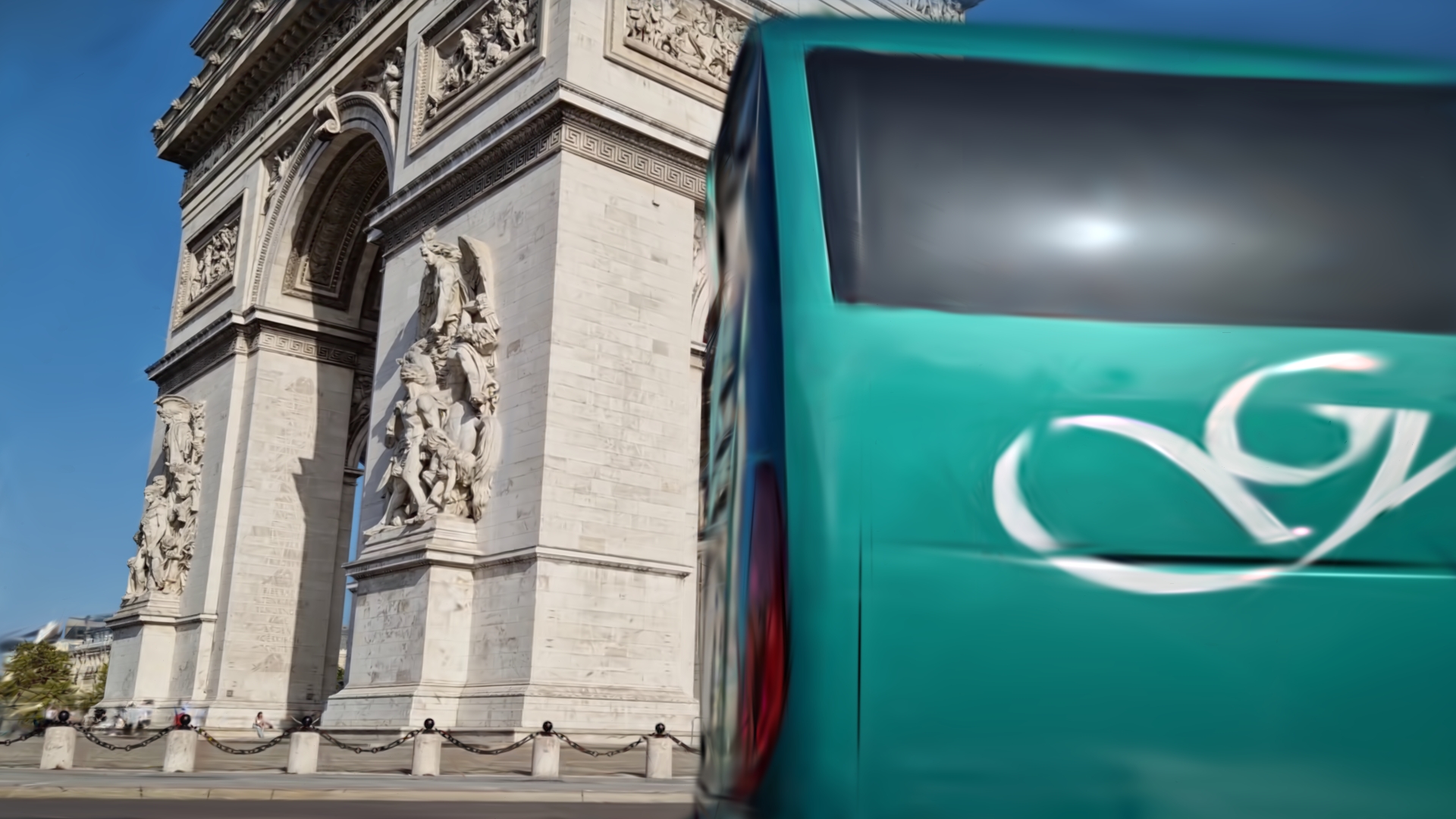} &
    \includegraphics[width=.3\linewidth]{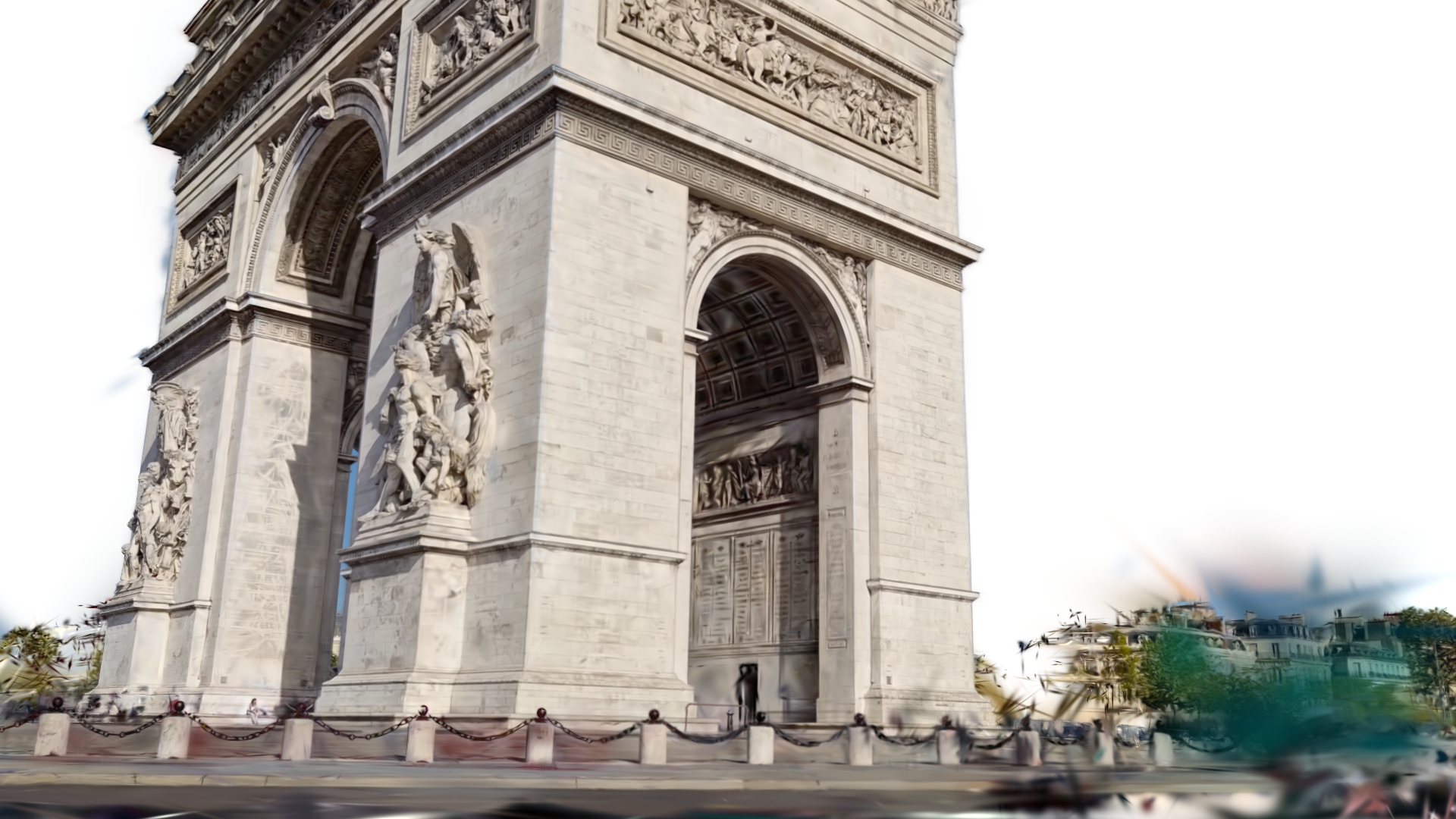}\\
    patio\_high &
    \includegraphics[width=.3\linewidth]{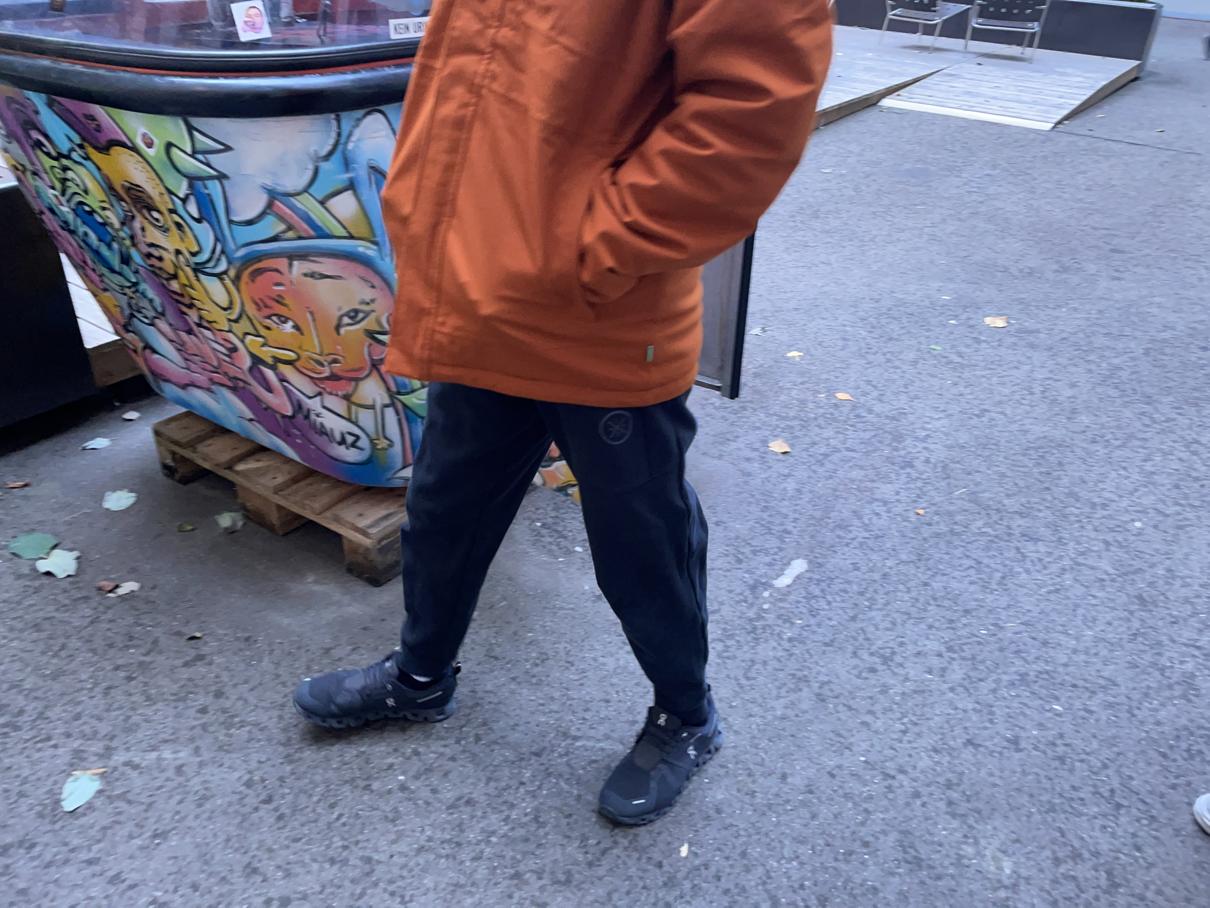} & 
    \includegraphics[width=.3\linewidth]{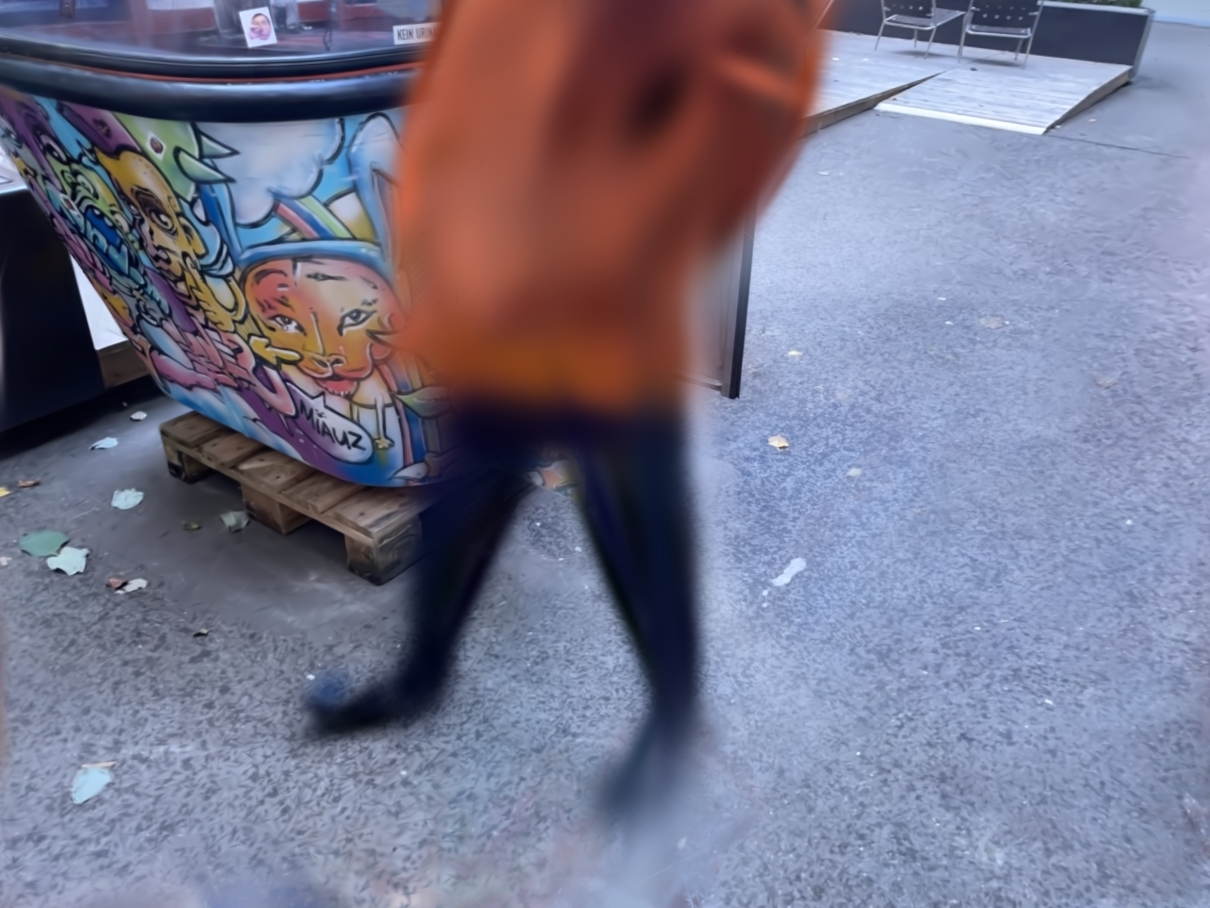} &
    \includegraphics[width=.3\linewidth]{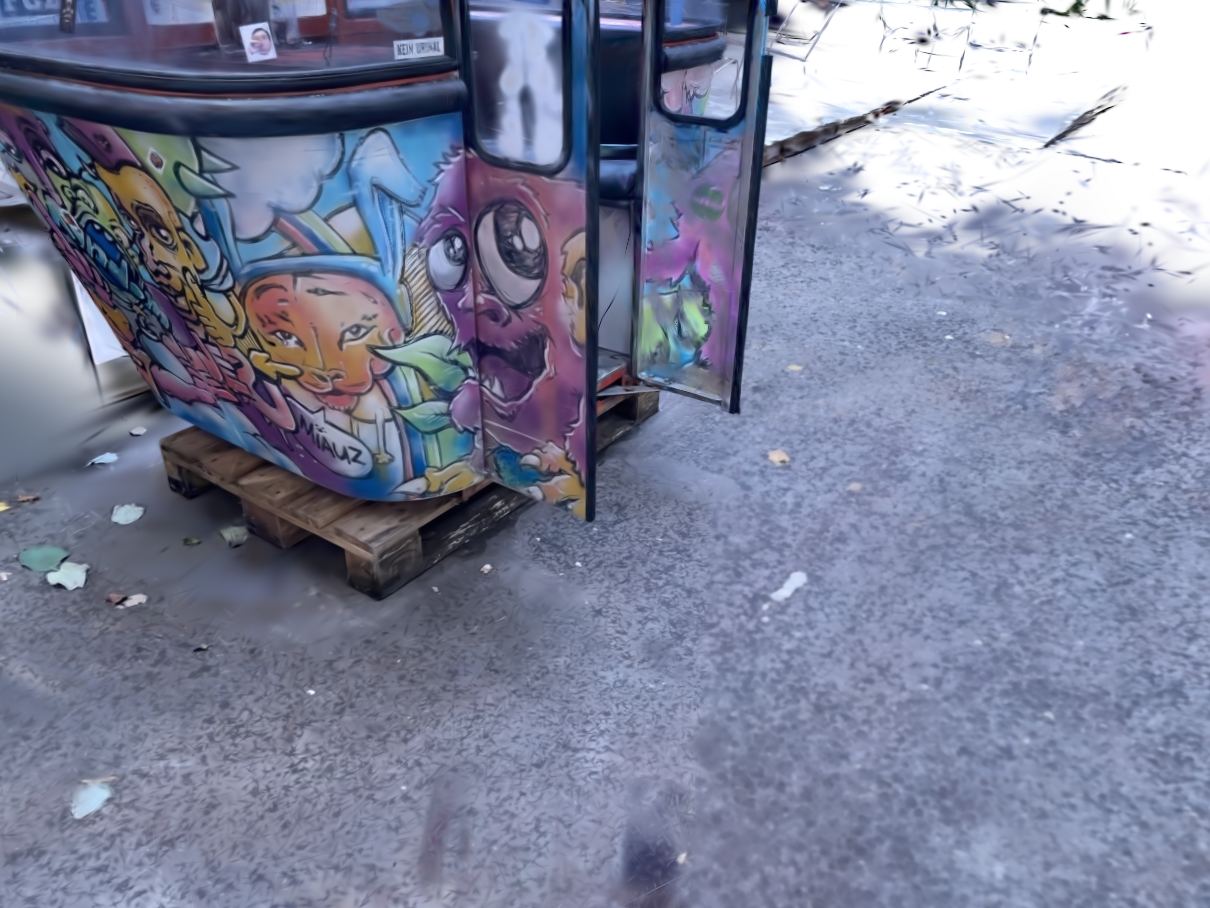}\\
    spot &
    \includegraphics[width=.3\linewidth]{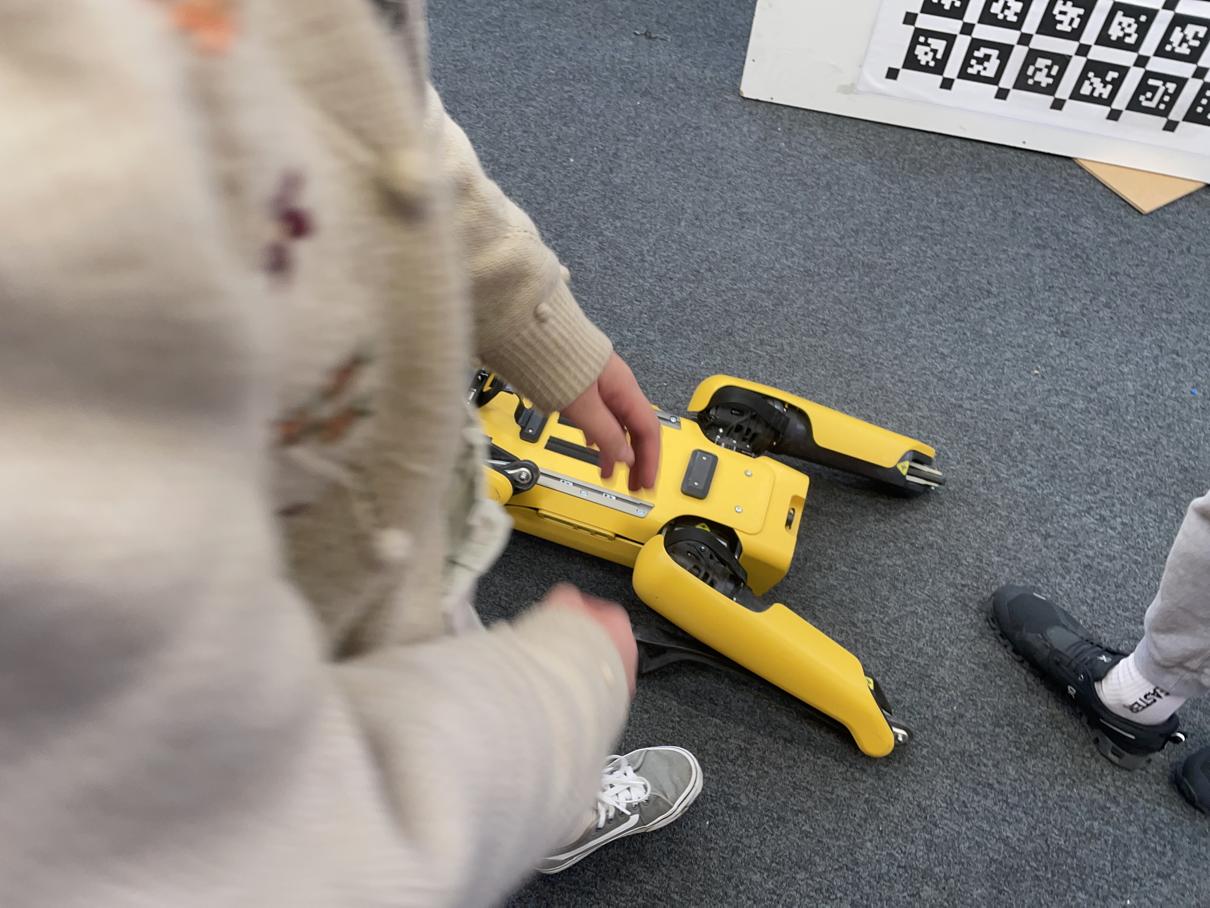} &
    \includegraphics[width=.3\linewidth]{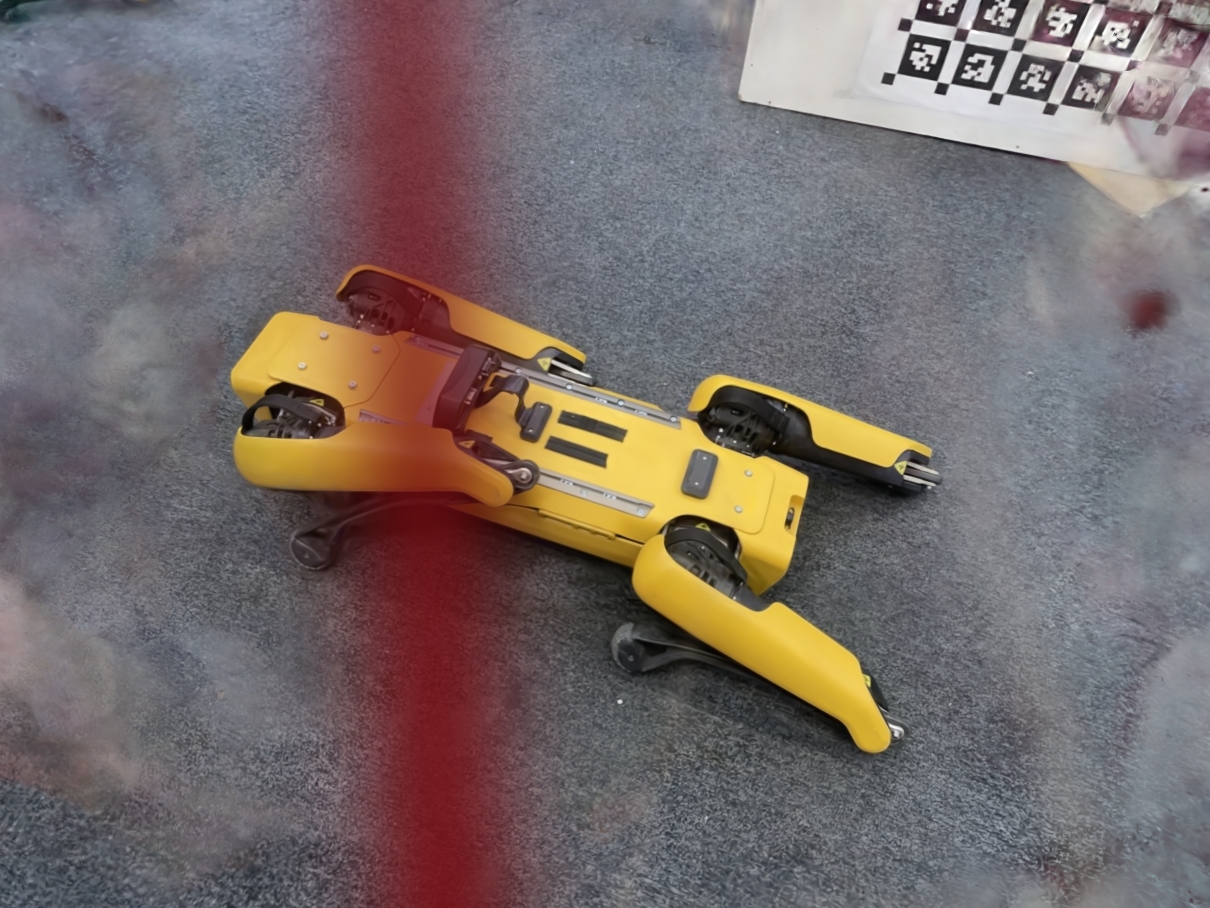} &
    \includegraphics[width=.3\linewidth]{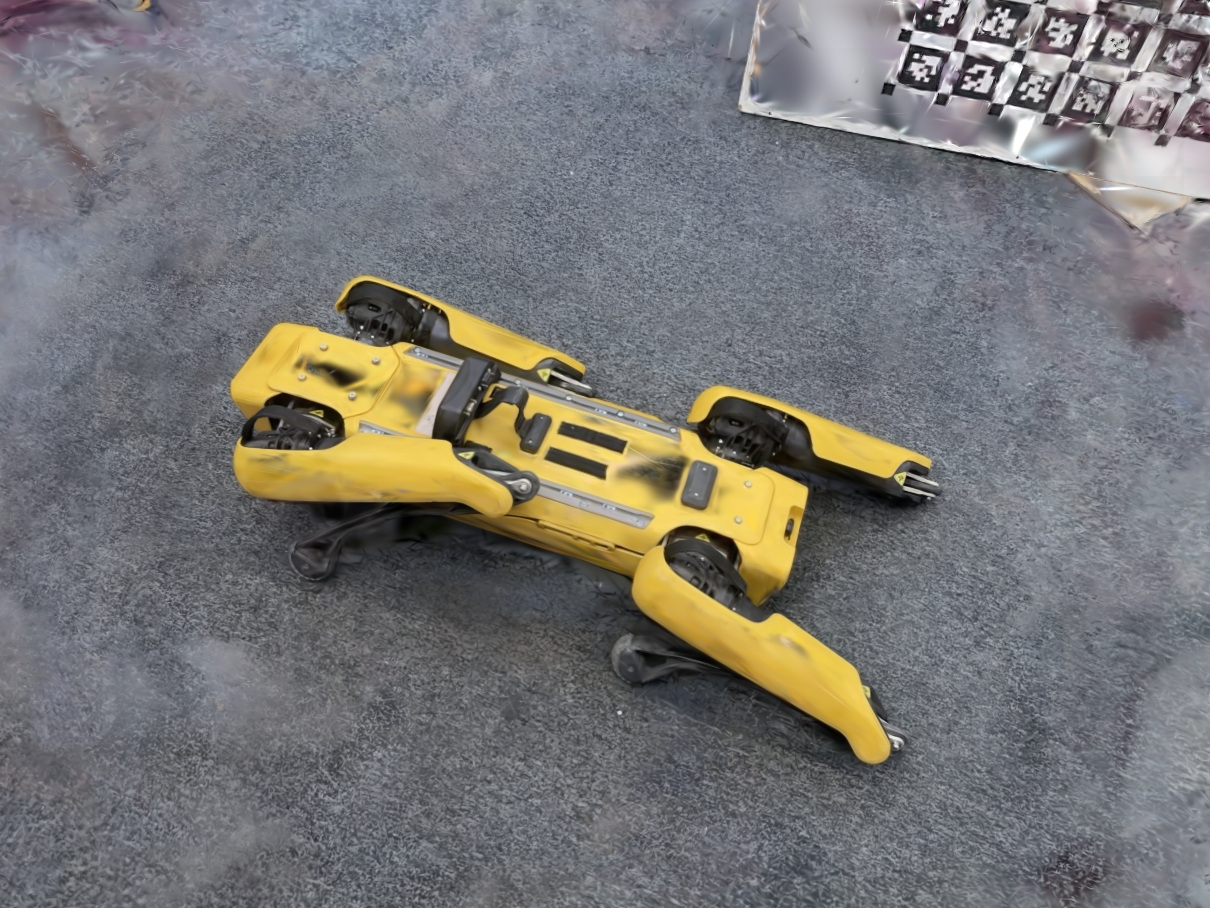}
  \end{tabular}
  \caption{NeRF-On-the-Go samples highlight our filter’s suppression of ambiguous regions from in-distribution camera poses.}
  \label{fig:unc}
\end{figure}

On NeRF On-the-go scenes, our rotation-only, per-intersection gradient behaves as a fast proxy for the Fisher Information Matrix. Instead of the usual expectation over Jacobian products, a decoupled single-sample gradient captures the dominant directional sensitivity (epistemic uncertainty) at render time. Large gradient magnitudes flag sparsely constrained or ambiguous regions; conditionally filtering the associated Gaussians suppresses anisotropic instabilities and reduces visible artifacts (see Fig.~\ref{fig:unc}).

\pagebreak
\section{Ablation}
We analyze the contribution of individual filtering components through a series of ablation studies.

\subsubsection{Single-Pass.} We first implement a single-pass approach in which Gaussians are not fully removed; instead, only their contributions are selectively filtered based on the intermediate gradient at the ray-Gaussian intersection point. As shown in Fig.~\ref{fig:single_pass_and_scale}(a), the centers of many Gaussians remain visible in the render. This occurs because the gradient magnitude is low near a Gaussian’s center. From Equation~\ref{eq:grad}, this behavior is expected: the $x_k$ term is near zero when the ray intersects the Gaussian near its mean. For the first Gaussian, there are no prior contributions along the ray, therefore the cumulative sensitivity remains at its initialized value of zero. As a result, initial Gaussians in a depth ordered list, which are often responsible for occluding distant geometry in extreme viewpoints, are preserved when intersected near their centers because their sensitivities are near zero.

\subsubsection{Scale-Incorporated.} 
\begin{figure*}[!h]
    \centering
    \begin{tabular}{ccc}
    \includegraphics[width=.25\linewidth]{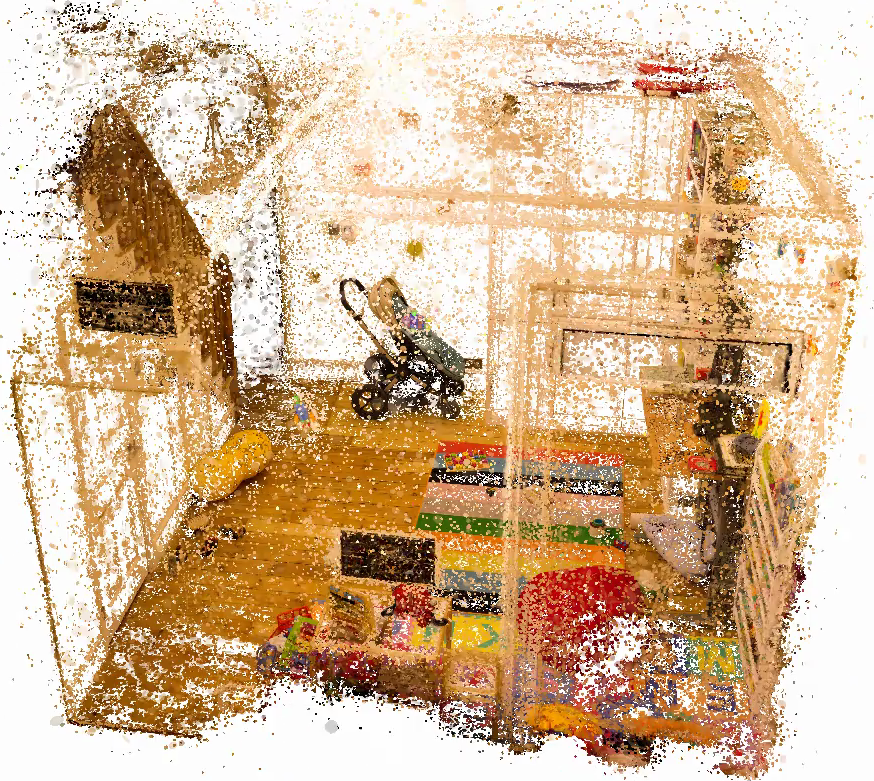} &
    \includegraphics[width=0.35\linewidth]{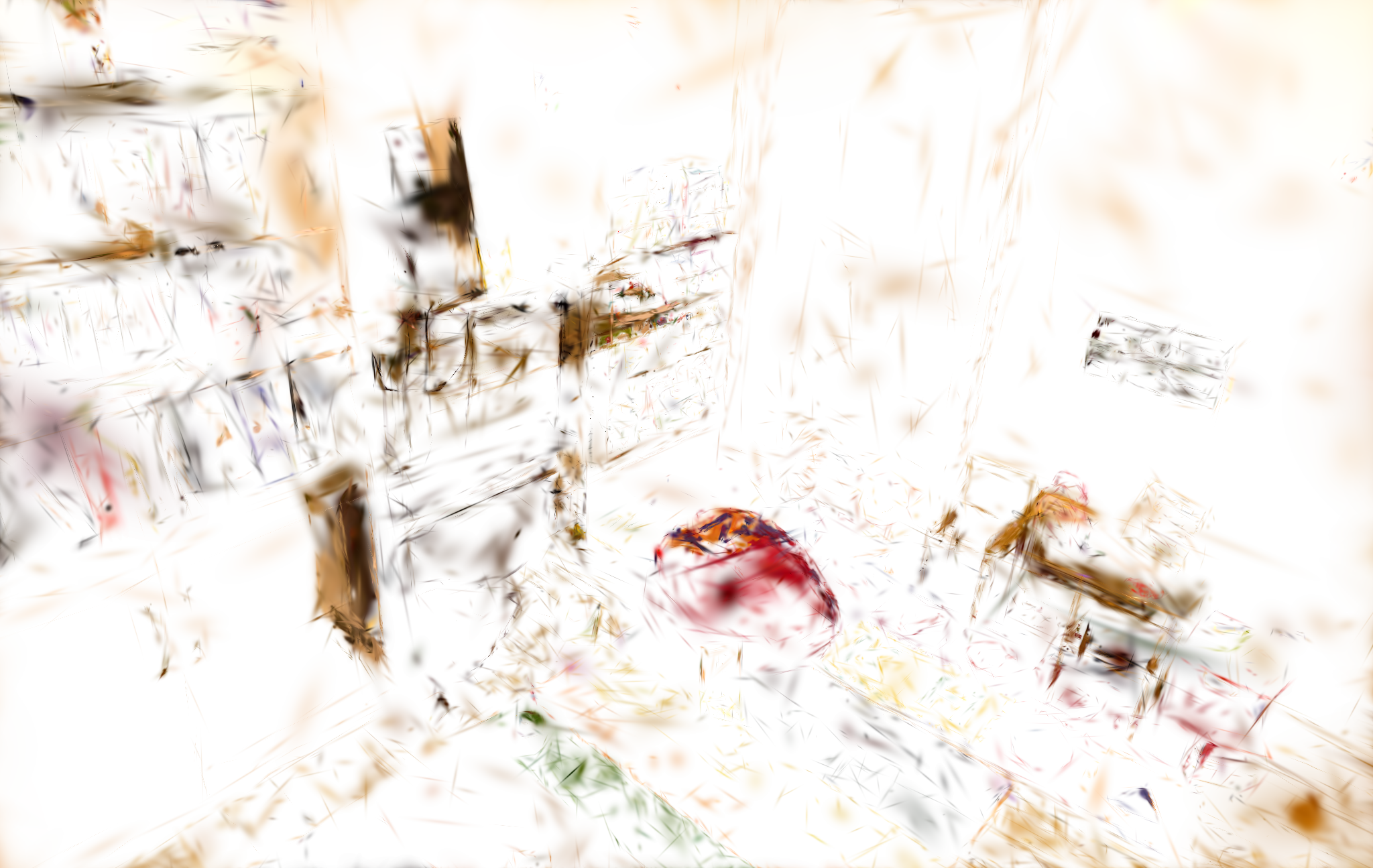}&
     \includegraphics[width=0.35\linewidth]{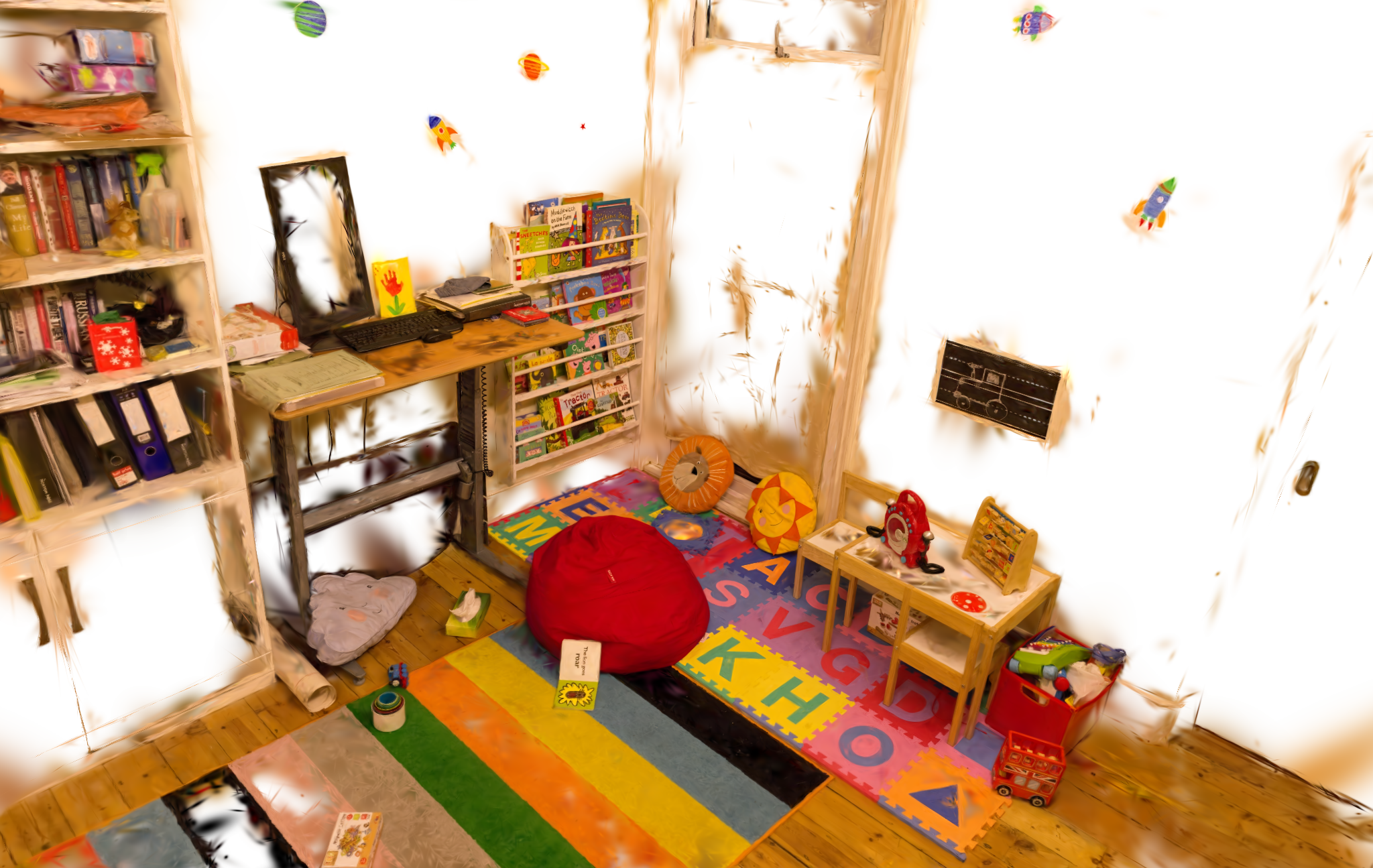}\\
     (a) & (b) & (c)
     \end{tabular}
    \caption{\\(a) Single-pass render of Playroom scene. $\tau_{grad.}=10^{-5}$\\
    (b) and (c) Loss of detail when including scale (b) vs. no scale (c)}
    \label{fig:single_pass_and_scale}
\end{figure*}

Next, we examine the impact of incorporating scale into the intermediate gradient calculation, as described in Section~\ref{sec43}.  This leads to a loss of anisotropic shape information, effectively normalizing Gaussians into unit spheres in their local coordinate space. Fig.~\ref{fig:single_pass_and_scale}(b) and (c) illustrate how this transformation causes elongated, noisy Gaussians to persist, while small Gaussians with scales less than 1 (which are critical for scene detail) are over-filtered.  This is due to the normalization process that increases the magnitude of the gradient vector which makes them more likely to exceed the rejection threshold.

\subsubsection{Filter Parameters.} 
\begin{figure*}[!h]
    \centering
    \begin{subfigure}[t]{.24\textwidth}
        \centering
        \includegraphics[width=\linewidth]{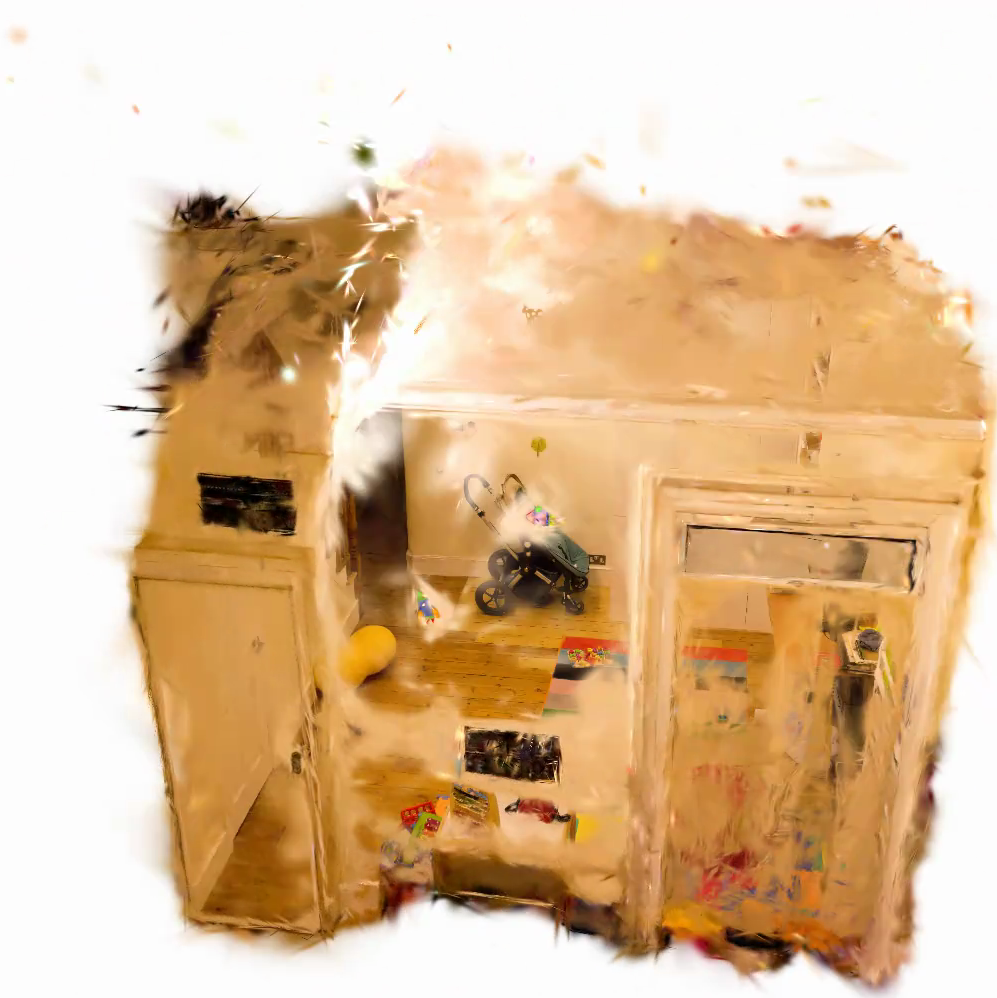}
        \subcaption{$\tau_{ratio}=0.5$, $\tau_{grad.}=0.0001$}
    \end{subfigure}\hfill
     \begin{subfigure}[t]{.24\textwidth}
        \centering
        \includegraphics[width=\linewidth]{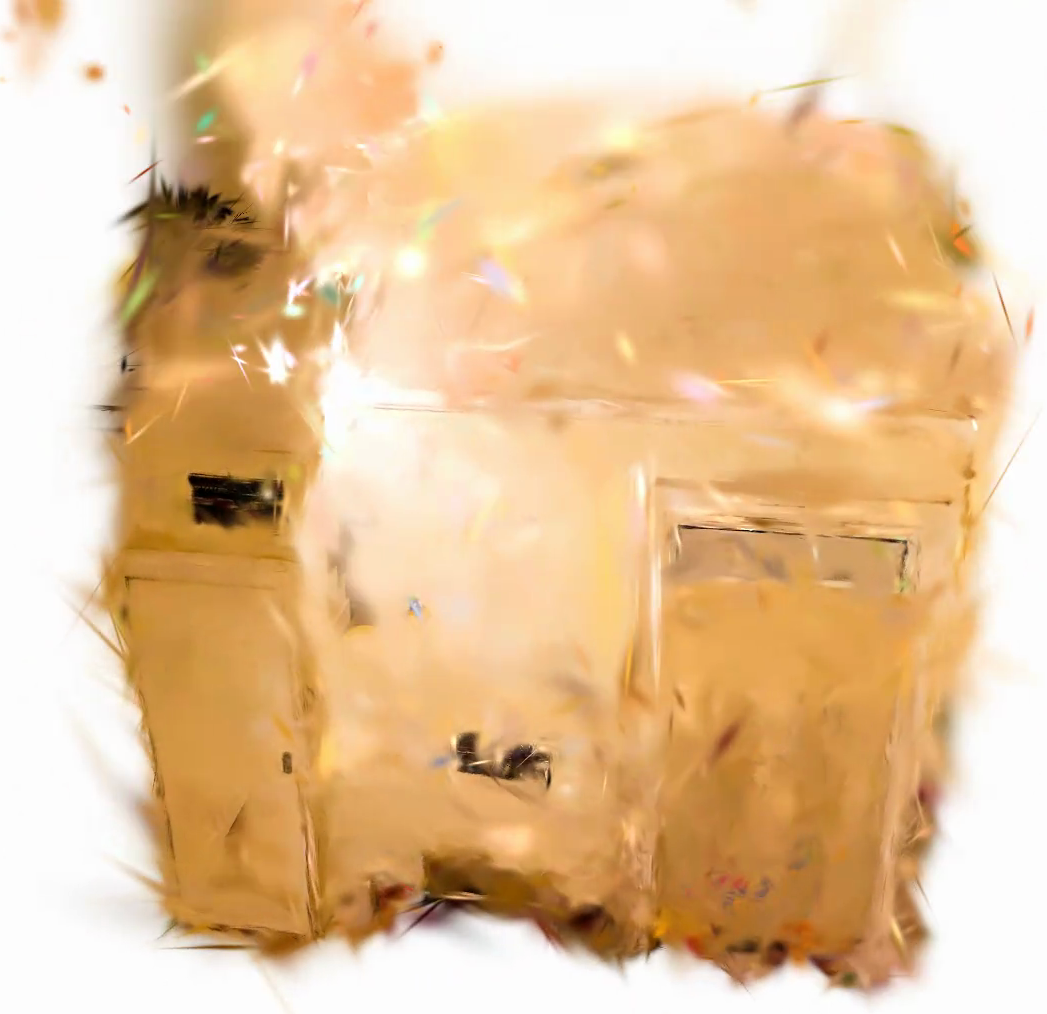}
        \subcaption{$\tau_{ratio}=0.5$, $\tau_{grad.}=0.0005$}
     \end{subfigure}\hfill
    \begin{subfigure}[t]{.24\textwidth}
        \centering
        \includegraphics[width=\linewidth]{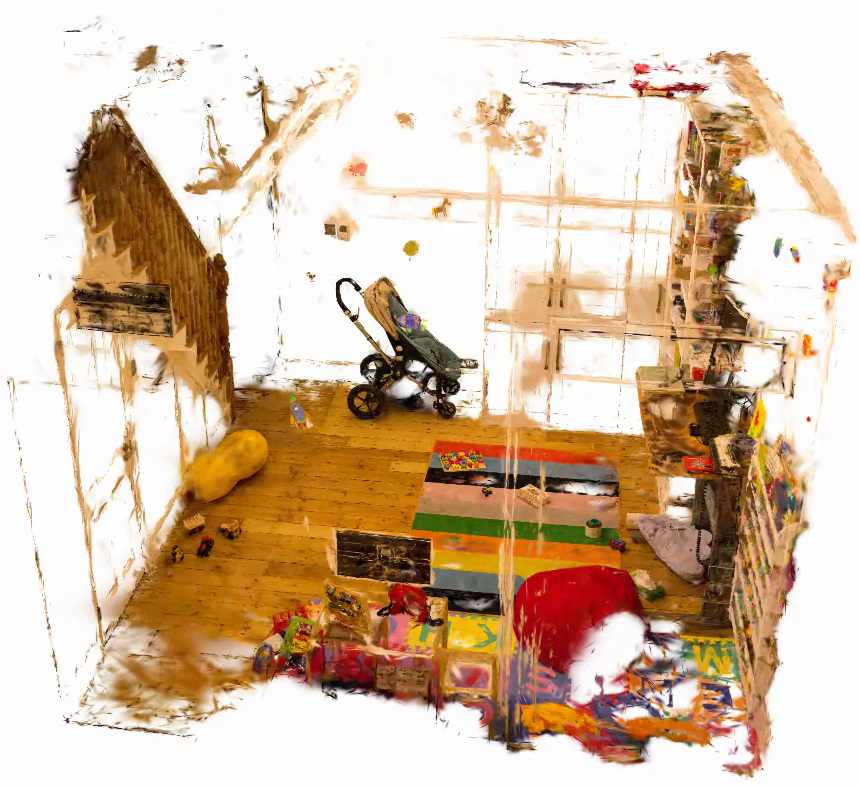}
        \subcaption{$\tau_{grad.}=10^{-5}$, $\tau_{ratio}=0.25$}
    \end{subfigure}\hfill
    \begin{subfigure}[t]{.24\textwidth}
        \centering
        \includegraphics[width=\linewidth]{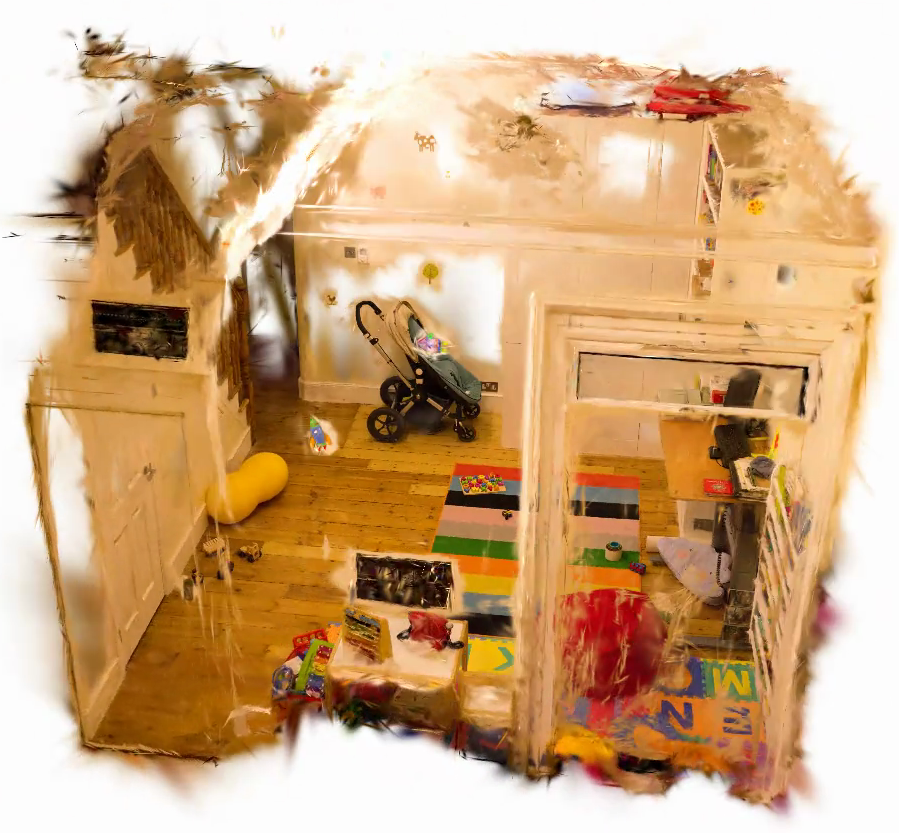}
        \subcaption{$\tau_{grad.}=10^{-5}$, $\tau_{ratio}=0.75$}
    \end{subfigure}
    \caption{Panels (a–b) hold $\tau_{ratio}=0.5$ and sweep $\tau_{grad.}$; panels (c–d) hold $\tau_{grad.}=10^{-5}$ and sweep $\tau_{ratio}$.}
    \label{fig:thresh_ablate}
\end{figure*}

The effect of the two parameters in isolation is shown in Fig. \ref{fig:thresh_ablate}. The subfigures (\ref{fig:thresh_ablate}.a and \ref{fig:thresh_ablate}.b) hold $\tau_{ratio}=0.5$ while varying $\tau_{grad.}$, and subfigures (c–d) hold $\tau_{grad.}=10^{-5}$ while varying $\tau_{ratio}$. Visual inspection suggests that using only one parameter yields different perceptual quality, whereas the best performance comes from a scene-specific combination of both tuned by the user.

\section{Conclusion}
We introduce a novel sensitivity measurement for 3DGS that identifies and filters anisotropic instabilities during rendering, without requiring retraining or scene-specific tuning. By analyzing intermediate gradient responses from the differentiable rasterization pipeline, our method targets the core source of generative uncertainty: Directional instability arising from anisotropic orientations. This filtering mechanism enables robust rendering even when users navigate freely beyond the original training views, a setting where standard 3DGS models often produce severe visual artifacts. Experimental results across complex, photorealistic datasets demonstrate consistent improvements in perceptual quality metrics surpassing baseline 3DGS and NeRF-based methods such as BayesRays.
\section*{Acknowledgement}
\vspace{-5pt}
This research was undertaken, in part, based on support from the Natural Sciences and Engineering Research Council of Canada Grants RGPIN-2021-03479 (NSERC DG) and ALLRP 571887 - 2021 (NSERC Alliance).
\bibliographystyle{splncs04}
\bibliography{references}
\end{document}